\documentclass[runningheads]{llncs}
\usepackage{graphicx}
\usepackage{subcaption}
\begin{document}
\title{Training dataset generation for automatic registration of a~duplicate bridge game}

\titlerunning{Training dataset generation for bridge game registration}
\author{Piotr Wzorek \and
Tomasz Kryjak \orcidID{0000-0001-6798-4444} }
\authorrunning{P. Wzorek, T. Kryjak}
\institute{AGH University of Science and Technology, \\ Al. Mickiewicza 30, 30-059 Krakow, Poland\\
\email{pwzorek@student.agh.edu.pl, tomasz.kryjak@agh.edu.pl} \\
}
%\institute{Princeton University, Princeton NJ 08544, USA \and
%Springer Heidelberg, Tiergartenstr. 17, 69121 Heidelberg, Germany
\maketitle              % typeset the header of the contribution
\begin{abstract}
This paper presents a~method for automatic generation of a~training dataset for a~deep convolutional neural network used for playing card detection. 
The solution allows to skip the time-consuming processes of manual image collecting and labelling recognised objects.
The YOLOv4 network trained on the generated dataset achieved an efficiency of 99.8\% in the cards detection task. 
The proposed method is a~part of a~project that aims to automate the process of broadcasting duplicate bridge competitions using a~vision system and neural networks. 

\end{abstract}
%
%
% --------------------------------------------------------------------------------------------------------------------------------------------------------------
\section{Introduction}

Duplicate bridge is a~popular logic game played with a~deck of 52 cards. 
Professional bridge tournaments are organised regularly, and the most prestigious ones are broadcasted live for hundreds of fans from around the world. 
Such broadcasts so far have been carried out manually by operators recording every play, which is a~monotonous and repetitive task. 
An attempt to automate this process using a~vision system is the motivation for the presented research. 

The key component of the considered application is the correct detection of cards and bidding calls, i.e., elements used during a~duplicate bridge game. 
We we have assumed the use of deep neural networks (DNN) to perform this task. 
The first step of a~DNN-based detector implementation is the preparation of a~training dataset. 
Using a~typical approach, this process is time-consuming and requires the collection of a~large number of images containing recognised objects. 
Then, each image must be labelled manually, i.e., each card has to be marked along with its class. 
The main contribution of this paper is the proposal of a~method for automatic dataset generation. 

The remainder of this paper is organised as follows. 
Section 2 discusses the issue of recording the course of a~duplicate bridge game. 
Section 3 presents the concept of card detection and bidding calls. 
Section 4 describes the proposed method for automatic generation of the training set. 
The paper concludes with a~summary and plans for further research work.

% --------------------------------------------------------------------------------------------------------------------------------------------------------------
\section{Registration of a~duplicate bridge game}

Bridge is a~logic game played by four players forming two pairs using classical 52 playing cards. 
Each player has thirteen cards at disposal. 
This game is characterised by a~very high level of complexity compared with other card games -- to succeed, players must have an excellent memory, logical and analytical skills, knowledge of probability calculus, and the ability to play in pairs.

\begin{figure}[!t]
\begin{subfigure}{0.47\textwidth}
\includegraphics[width=\textwidth]{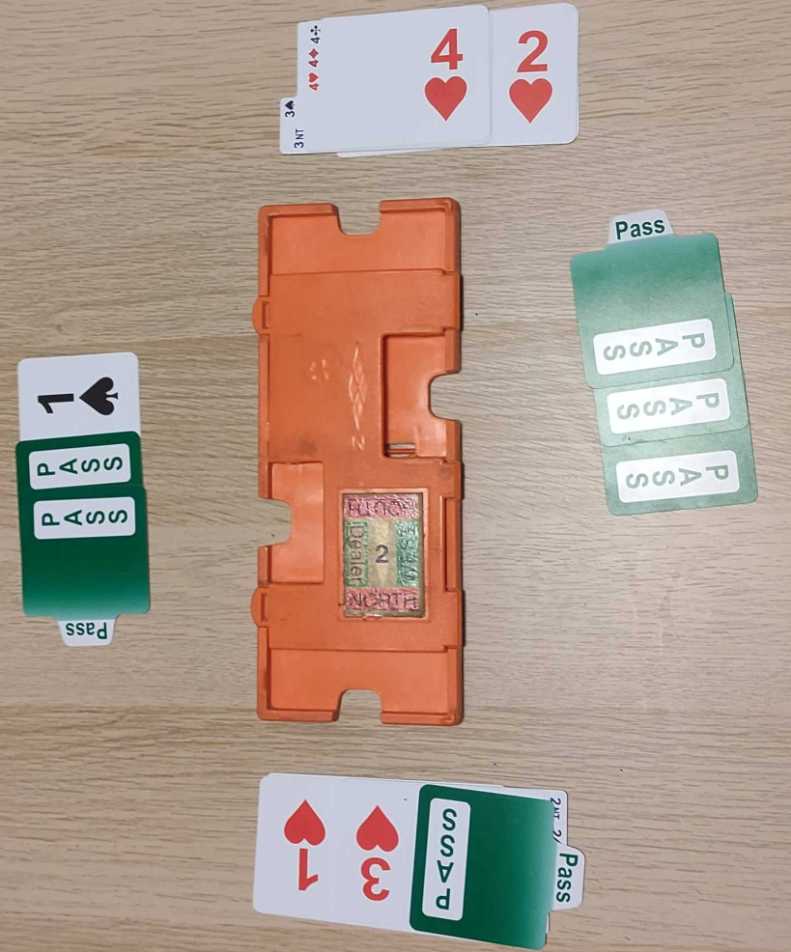}
\caption{Example auction.} 
\label{fig1a}
\end{subfigure}
\begin{subfigure}{0.5\textwidth}
\includegraphics[width=0.97\textwidth]{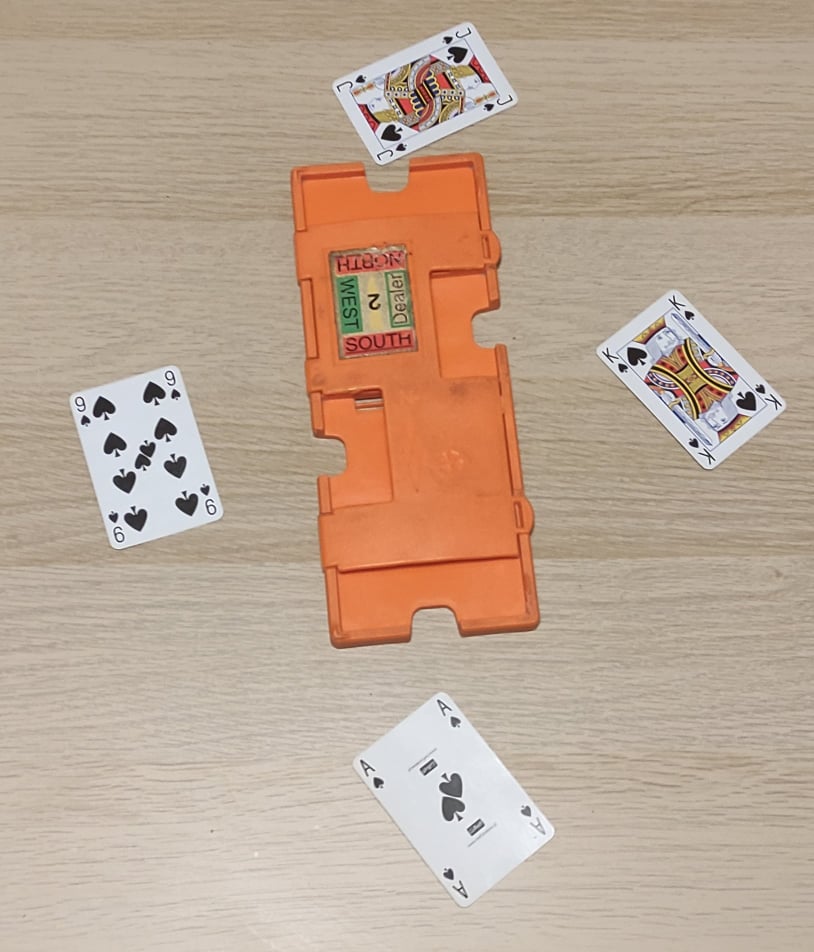}
\caption{Example trick.} 
\label{fig1b}
\end{subfigure}
\caption{The course of a~bridge game: a) the picture shows example bidding calls -- cardboard markers with symbols, b) each player has played one card, and the trick is won by the player who played the ace of spades.} 
%\label{fig1}
\end{figure}

A~bridge game consists of two consecutive phases: auction and play. 
The first phase is an exchange of information about the players' cards. 
For this purpose, the players use the so-called bidding calls, i.e., cardboard markers with symbols that are defined in the game rules (Fig. \ref{fig1a}). 
In the second phase of the game, the players play one card in turn. 
This is called a~trick (Fig. \ref{fig1b}) and the winner is the person who played the highest card. 
The number of tricks each pair must take is determined by the auction. 
The play consists of 13 tricks -- the game ends when each player has played all cards.

\begin{figure}[!t]
\includegraphics[width=\textwidth]{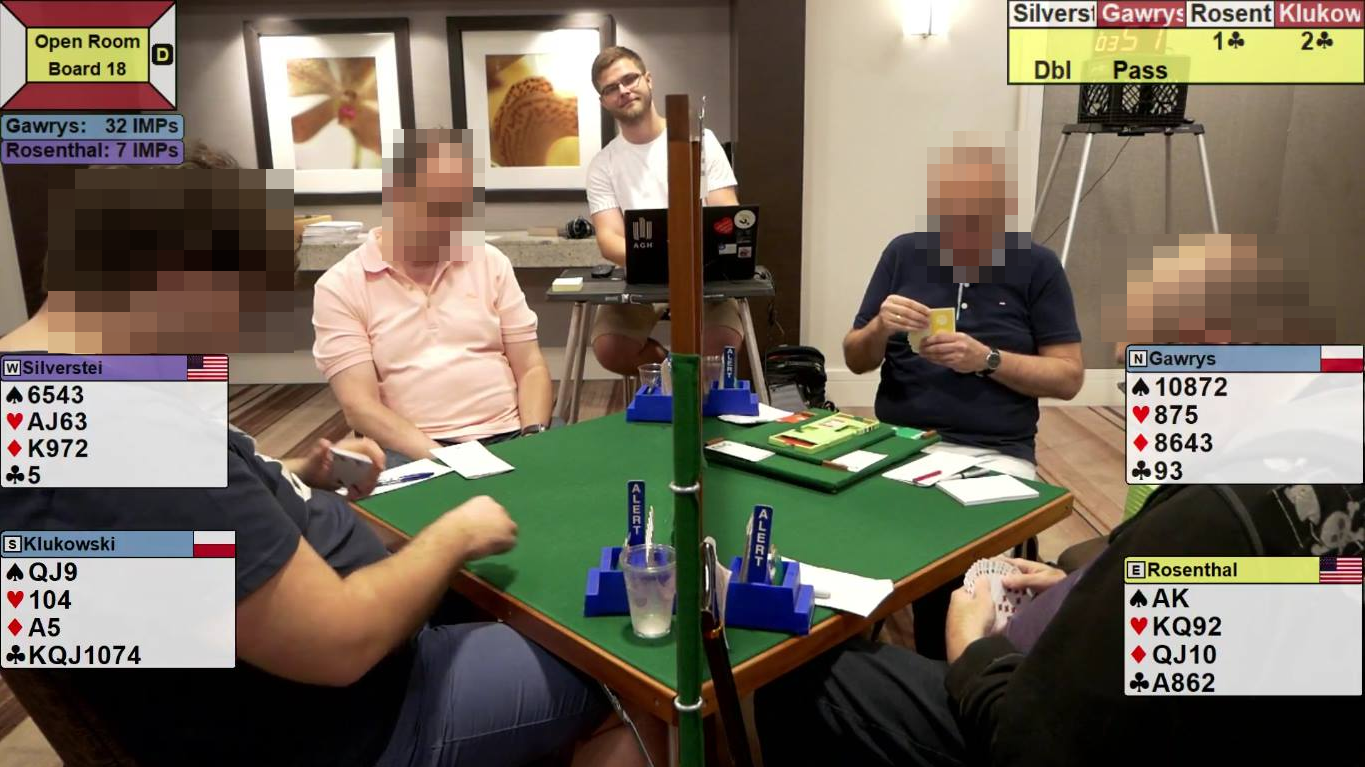}
\caption{Broadcast of the Spingold tournament played in August 2018 in Atlanta, Georgia. The screen shows the cards of each player and the auction process (upper right corner). In the centre, the broadcast operator can be seen recording each of the plays.} 
\label{fig2}
\end{figure}

\begin{figure}[!t]
\includegraphics[width=\textwidth]{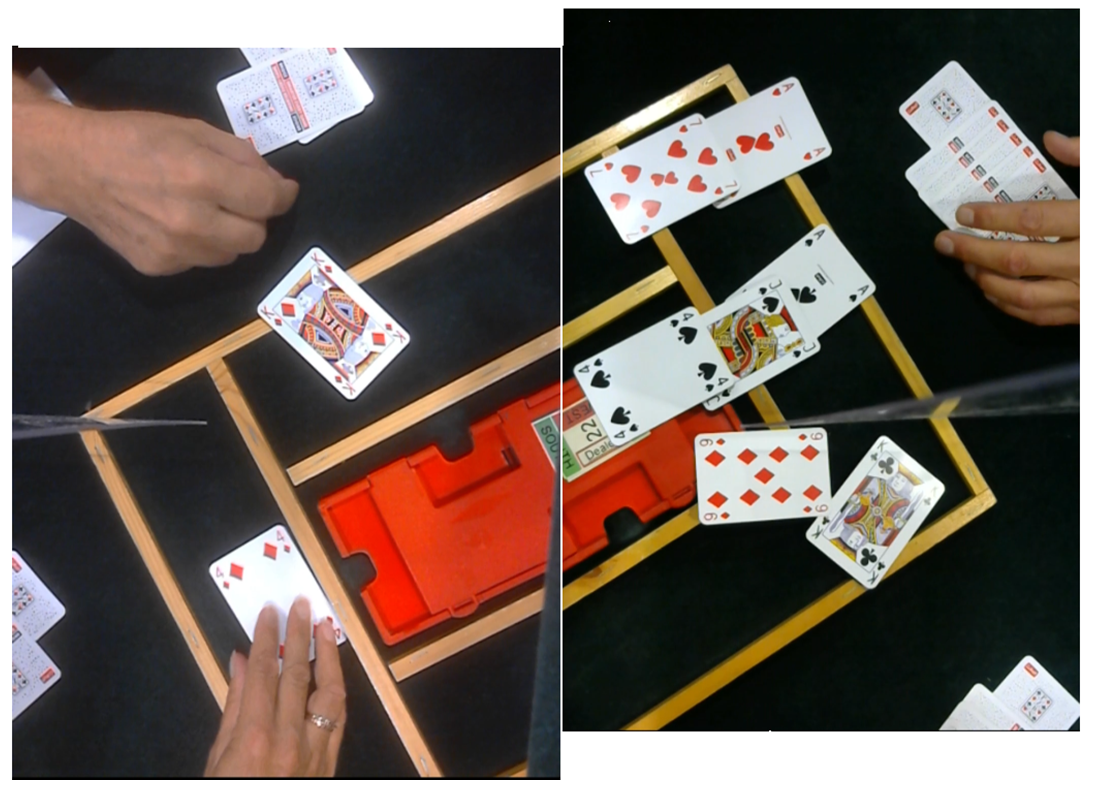}
\caption{The course of a~bridge game recorded by two cameras placed on the wooden screen.} 
\label{fig3}
\end{figure}

At the most prestigious tournaments, a~live broadcast is organised. 
Operators take their places at the tables selected for broadcast and record every bidding call and every card played (Fig. \ref{fig2}). 
The subject of this research is an attempt to automate this process using a~vision system. 
For this purpose, a~wooden screen, which is placed along the diagonal of the bridge table, is used (its purpose is to reduce the chance of passing unauthorised information to one's partner). 
Mounting two cameras on this screen, one for each half of the table, allows recording all objects on the table (Fig. \ref{fig3}). 
On the basis of correct detection of the cards played and bidding calls, using knowledge about the rules of the game, it is possible to reconstruct its course.

% --------------------------------------------------------------------------------------------------------------------------------------------------------------
\section{Detection of cards and bidding calls}

In the considered task, 90 object classes should be detected by the neural network: 52 different cards and 38 unique bidding calls.
The input to such a~detector is a~single video frame recorded by a~camera placed on the screen. 
The output is a~list of objects in the image, along with the information about their location, size, and class membership.
The key element to achieve high detection performance is the preparation of a~suitable training dataset. 
It should contain many different versions of the recognised objects to obtain a~neural network that works in a wide range of conditions.

In the first stage of the study, we have analysed the complications related to the detection of cards and bidding calls used during a~bridge game. 
For this purpose, we have gathered and studied video recordings of duplicate bridge games. 
We have identified the following potential problems:

\begin{enumerate}
    \item Plastic cards and bidding calls are made of a~very reflective material, which in unfavourable lightning conditions causes difficulties in their proper classification. The dataset should therefore contain images of cards illuminated at various angles, with varying light intensity and colour.
    \item Bridge players do not always play cards by placing them on the table -- some hold them in their hand, which results in partial occlusions. 
    Also, according to the rules of bridge, one player places all 13 cards on the table during the play phase in such a~way that some of them are only partially visible (Figure \ref{fig7}). Therefore, the dataset should contain pictures of cards and calls with varying degrees of coverage, at multiple angles, placed in different places on the table.
    \item Any centre-symmetrical playing cards are allowed during bridge competitions. 
    Therefore, some significant differences between objects of the same class may occur -- so the cards and calls of different patterns and shapes are used. 
    The dataset should include at least 2 different patterns of cards and bidding calls with the possibility of an easy extension.
    \item In the considered case, objects of different classes hardly differ from one another, i.e., playing cards are rather similar. Therefore, the network must be able to distinguish between each card and call, even though they differ only in the numbers and symbols drawn on them. The dataset should be relatively large to make it easy for the network to distinguish between each class of objects.
\end{enumerate}

Nowadays, object detection methods with the highest accuracy are based on deep convolutional neural networks \cite{1}. 
For the described application, from a~functional point of view, high detection efficiency and relatively low latency are essential -- to enable real-time transmission of the bridge game, the detector must be characterised by a~high operating speed. 
Additionally, it is desirable that the computations can be performed on a~typical laptop computer or even an embedded hardware platform such as Nvidia Jetson series chip. 
Therefore, we have used the YOLOv4 network \cite{2} which has very good performance in both metrics and a relatively low computational complexity, for the presented problem.

% --------------------------------------------------------------------------------------------------------------------------------------------------------------
\section{Training dataset generation}

The classical approach to preparing a~training dataset involves collecting and labelling images containing objects that the trained network is supposed to recognise, which is a~monotonous and time-consuming process. 
Therefore, we proposed a~method to generate the dataset in a~mostly automatic way. 
This approach was inspired by the ``Playing Card Detection'' project, which was made publicly available in the GitHub repository and documented in a~video on YouTube \cite{3}. 
Based on the analysis of this solution, we developed our own implementation of the automatic dataset generation tool. 
The process of preparing the set for cards and bidding calls is analogous, however, in this paper we address only the generation of the dataset for the card detection network.

\begin{figure}[!t]
\includegraphics[width=\textwidth]{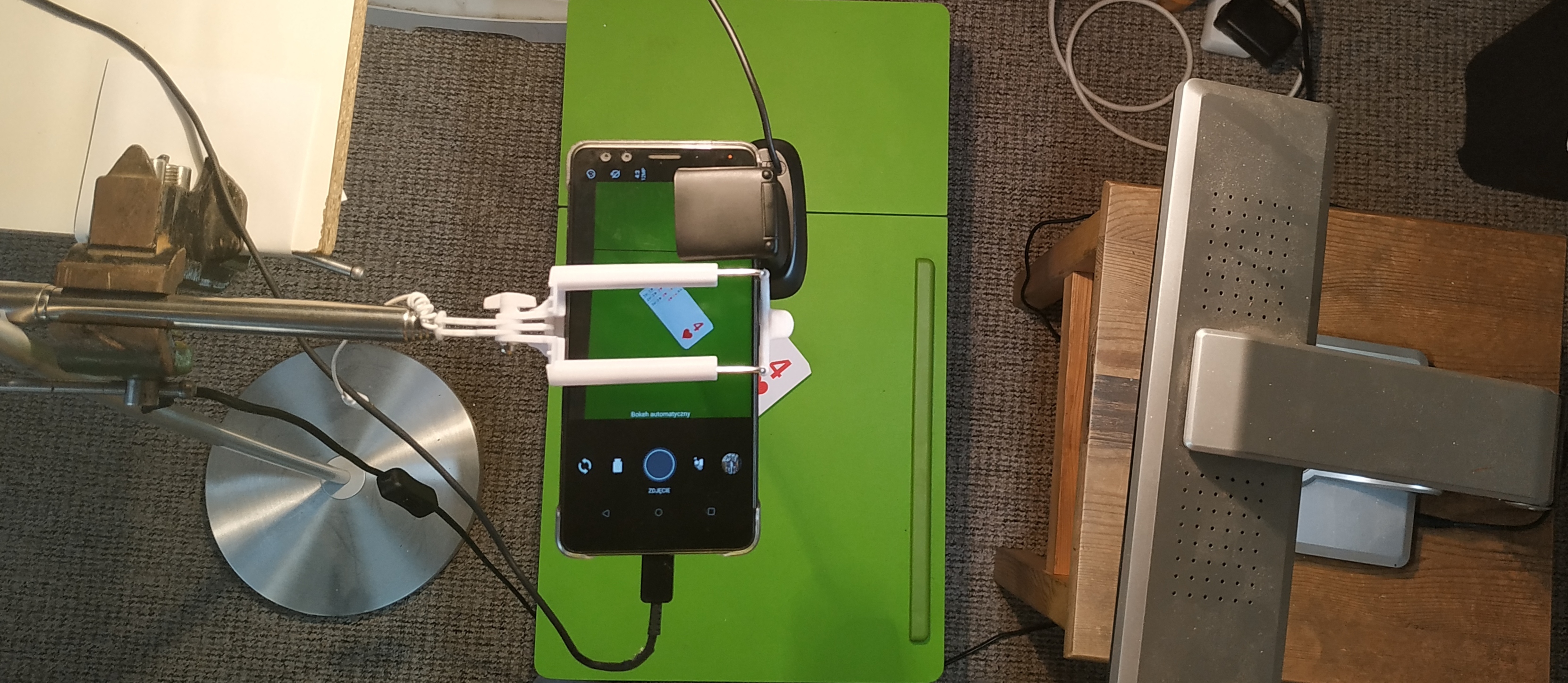}
\caption{Video preparation for each recognised object.} 
\label{fig4}
\end{figure}

\begin{figure}[!t]
\begin{center}
\includegraphics[width=0.7\textwidth]{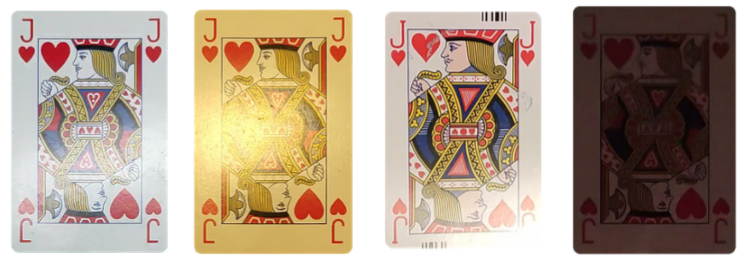}
\end{center}
\caption{Photos of the jack of hearts illuminated in different ways. Two different patterns were used.} 
\label{fig5}
\end{figure}

In the first step of the dataset preparation, we recorded 30-second videos in which we placed a~single card on a~uniform-colour background. 
During the video, we illuminated the card from different angles, with varying light intensity and colour (Fig. \ref{fig4}). 
We used for this purpose two different lamps, varying their position relative to the cards and adjusting the illumination intensity. 
We decided to use real cards, as we assumed that achieving a~similar effect in a~graphics software would require much more work and time.

\begin{figure}[!t]
\begin{center}
\includegraphics[width=0.7\textwidth]{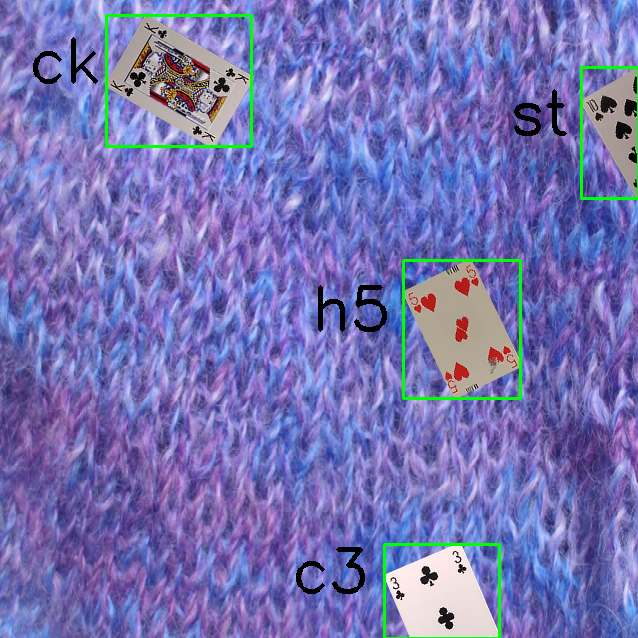}
\end{center}
\caption{An example image from the card dataset with the labels highlighted. The green rectangles indicate the location of the object and the two-character description is the name of the class. The first character gives information about the colour of the card (C -- clubs, D -- diamonds, H -- hearts, S -- spades), and the second character gives information about its rank.} 
\label{fig6}
\end{figure}

Then we manually marked the cards in the first frame of each sequence. 
We used the mask created this way for the remaining frames (the card did not change its position). 
We used a~simple code in Python and the OpenCV library \cite{4} to implement this step. 
Finally, we obtained a~set of images of single cards illuminated at different angles, with light of varying colour and intensity (Figure~\ref{fig5}).

\begin{figure}[!t]
\begin{center}
\includegraphics[width=0.7\textwidth]{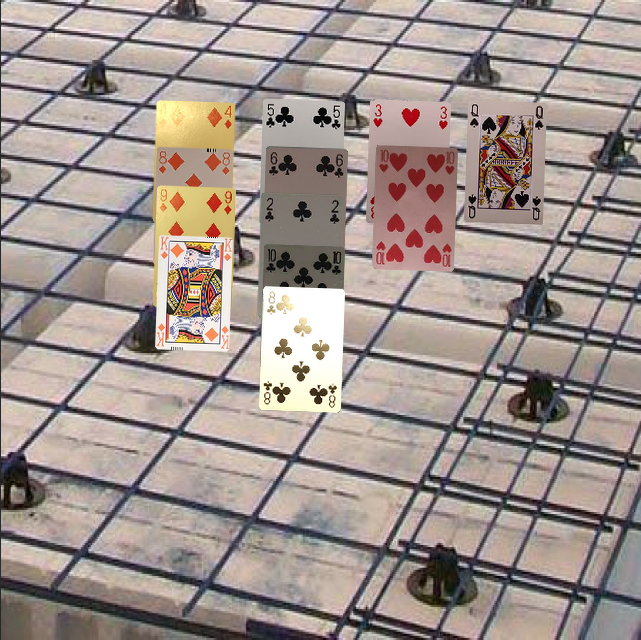}
\end{center}
\caption{Sample card dataset image -- ``bridge dummy''.} 
\label{fig7}
\end{figure}

In the next step, we placed the prepared cards on a~selected background. 
For this purpose, we used the Describable Textures Dataset \cite{5}, which contains 5640 texture images. 
For each image, we randomly drew the number of cards to be placed, the position of each card, and the angle at which they were rotated. 
At the same time, we recorded the position and class of each card. 
Consequently, using the proposed application, it is possible to generate datasets of any size. 
Moreover, there is no need to label each of the images manually -- the position stored for each card can be used to generate labels. 
An example image from a~dataset prepared in this way is shown in Figure \ref{fig6}.

We supplemented the dataset with pictures typical for the game of bridge.  
A~special case is the so-called ``bridge dummy'' -- the cards of one of the players laid out on the table. 
In the case of dummy detection, the task is much more complicated -- there is a~large number of partially covered cards on a~small area of the table. 
To facilitate the network's correct recognition of objects in such situations, we added a~function that generates cards in this specific arrangement -- sorted by colour and placed on the table (Fig. \ref{fig7}). 
For this purpose, we limited the allowed angle of rotation of the cards and fixed their position relative to each other.

% --------------------------------------------------------------------------------------------------------------------------------------------------------------
\section{Summary}

The proposed method of automatic dataset generation for cards detection and classification makes it possible to obtain any number of images of any size, which can be used to train a~deep convolutional neural network. Moreover, based on the preliminary analysis of the considered detection problem, it is possible to take into account a~number of complications in the process of generating the training set -- variable illumination, position, as well as characteristic card layouts. 

We used the dataset generated by the above-described method, consisting of nearly 32000 images to train the YOLOv4 network and, according to preliminary experiments, the obtained detector achieved an 99.8\% accuracy on the test set (nearly 8000 images). 
We will use it for further stages of the work on the automatic registration of the course of a~duplicate bridge game. It is worth noting that the generation process of the bidding calls dataset can be conducted in an analogous way. 
Moreover, the described method could be used for a~number of different applications. 

This research was supported by the AGH University of Science and Technology project no. 16.16.120.773 and
PLGrid Infrastructure.


\end{document}